# Improving O-RADS Risk Stratification from Ultrasound Reports: A Comparative Evaluation of Hybrid versus End-to-End LLM Reasoning Strategies

Xiaotong Tan[1,2#], Chunli Qiu[3#], Xin Liu[3], Qing Huang[1,2], Guangli Zhou[4], Bo Gao[5], Xiaoyan Song[6], Shuyan Wang[7], Xiuqin Wang[8], Wufeng Xue[1,2], Ruobing Huang[1,2], Dong Ni[2,9,10,11], Guowei Tao[3,12]*, Jun Cheng[1,2]*

1. National-Regional Key Technology Engineering Laboratory for Medical Ultrasound, Guangdong Key Laboratory for Biomedical Measurements and Ultrasound Imaging, School of Biomedical Engineering, Shenzhen University Medical School, Shenzhen University, Shenzhen 518055, China

2. Marshall Laboratory of Biomedical Engineering, Shenzhen University, Shenzhen 518055, China

3. Department of Ultrasound, Qilu Hospital of Shandong University, Jinan 250012, China

4. Department of Ultrasound, Tengzhou Central People's Hospital, Tengzhou 277500, China

5. Department of Ultrasound, Zibo Central Hospital, Zibo 255036, China

6. Department of Ultrasound, Shengli Oilfield Central Hospital, Dongying 257034, China

7. Department of Ultrasound, Sunshine Union Hospital, Weifang 261061, China

8. Department of Ultrasound, Taierzhuang District People’s Hospital, Zaozhuang 277400, China

9. School of Artificial Intelligence, Shenzhen University, Shenzhen 518060, China

10. National Engineering Laboratory for Big Data System Computing Technology, Shenzhen University, Shenzhen 518060, China

11. School of Biomedical Engineering and Informatics, Nanjing Medical University, Nanjing 211166, China

12. Shandong Key Laboratory of Reproductive Health and Birth Defects Prevention and Control, Jinan 250012, China

# Co-first authors

* Co-corresponding authors

**Corresponding authors:**

Guowei Tao, MD

107 Wenhua W Rd, Department of Ultrasound, Qilu Hospital of Shandong University, Jinan 250012, China

Email: taoguowei2006@163.com

Jun Cheng, PhD

1066 Xueyuan Rd, School of Biomedical Engineering, Shenzhen University Medical School, Shenzhen University, Shenzhen, 518055, China

Email: chengjun583@qq.com

## Abstract

**Background:** Automating clinical guideline-based decision-making with large language models (LLMs) remains challenging because of reliability, hallucination, and limited interpretability. We compared the performance of LLMs and reasoning strategies for automated Ovarian-Adnexal Reporting and Data System (O-RADS) classification from free-text pelvic ultrasound reports.

**Methods:** In this retrospective study, consecutive patients with ovarian masses who underwent pelvic ultrasound were included. Eight LLMs were tested with three reasoning strategies: implicit-knowledge end-to-end, rule-informed end-to-end, and a feature-based hybrid architecture that decoupled feature extraction from rule-based classification. The reference standard was O-RADS categorization established by expert consensus.

**Results:** A total of 310 women with 390 ovarian masses were evaluated. The feature-based hybrid architecture using Gemini 3.6 Flash demonstrated the best performance, achieving an accuracy of 99.2% (387 of 390) and almost perfect agreement with the reference standard (weighted kappa = 1.00; 95% CI: 0.99-1.00). Its performance surpassed that of original clinical reports (accuracy, 87.7% [342 of 390]; weighted kappa = 0.94; 95% CI: 0.91-0.96) and end-to-end LLM strategies (accuracy range, 65.6% [256 of 390] to 95.9% [374 of 390]). For structured feature extraction, Gemini 3.6 Flash demonstrated higher overall accuracy than Claude Fable 5 (98.9% vs 97.8%; P < 0.001). The hybrid architecture reduced misclassification errors and mitigated the overstaging tendency observed in original reports.

**Conclusion:** The feature-based hybrid LLM architecture that separates clinical feature extraction from deterministic guideline execution enables highly accurate, reliable, and interpretable automated O-RADS classification, providing a promising approach for standardized, guideline-based clinical decision-making.

## 1. Background

Accurate preoperative risk assessment of ovarian adnexal masses is essential for guiding appropriate clinical management and improving patient outcomes [1, 2]. Ultrasound (US) is the first-line imaging modality for adnexal evaluation because of its safety, cost-effectiveness, and wide availability. To standardize risk stratification based on sonographic features, several structured systems have been proposed, including the IOTA Simple Rules [3], the ADNEX model [4], and the Ovarian-Adnexal Reporting and Data System (O-RADS) [5, 6].

Among these systems, O-RADS has been adopted in clinical practice due to its explicit risk stratification (categories 0–5) and corresponding management recommendations. Nevertheless, its effective implementation relies on accurate interpretation of complex rules. In routine practice, integrating multiple morphological descriptors—such as lesion architecture, solid components, and vascularity—into the O-RADS framework remains challenging, particularly for less experienced ultrasound practitioners [7, 8]. Although auxiliary tools, such as the ACR Guidance App, provide point-of-care decision support, frequent manual consultation may disrupt clinical workflow and limit scalability, underscoring the need for automated and intelligent solutions.

Recent advances in large language models (LLMs) have demonstrated substantial potential in medical natural language processing tasks [9–12]. However, direct end-to-end guideline-based classification by LLMs remains prone to instability and hallucinations, limiting their reliability in clinical tasks [13–18]. To address these limitations, prior studies have explored enhanced reasoning strategies, including Chain-of-Thought prompting [19–21] and two-stage architectures [22–24]. Among these approaches, decoupling separates clinical information extraction from guideline-based decision execution: the LLM first converts unstructured clinical text into structured variables, which are subsequently mapped to final categories using deterministic rule-based algorithms. This design leverages the strong language understanding capability of LLMs while maintaining the reproducibility and transparency of explicit clinical rules. By enabling reliable extraction of structured information from complex clinical narratives [25–28], decoupled frameworks provide a promising solution for guideline-based clinical workflows, such as O-RADS classification.

While recent studies have investigated LLM-based approaches for automated O-RADS MRI grading [23], applying these models to O-RADS US poses distinct challenges. Compared with MRI reports, ultrasound reports frequently exhibit significant heterogeneity and non-standardized free-text narratives. Therefore, this study aims to systematically evaluate and optimize LLM-based strategies for automated O-RADS US classification. By addressing the challenges of parsing unstructured ultrasound reports, we seek to provide an efficient tool that enhances risk assessment consistency and supports standardized clinical decision-making.

## 2. Methods

### 2.1. Study Design

To systematically evaluate LLM performance in O-RADS ultrasound classification, we designed three distinct reasoning strategies to assign O-RADS categories from unstructured reports (Fig. 1). For all strategies, few-shot prompting was applied by incorporating two representative examples.

**Implicit-knowledge end-to-end strategy:** This strategy relied solely on the model’s internal parametric knowledge for end-to-end inference. Prompts included task instructions, few-shot examples, and textual descriptions of common ultrasound features, while explicit O-RADS scoring rules and decision logic were excluded.

**Rule-informed end-to-end strategy:** To mitigate potential knowledge gaps, this strategy explicitly incorporated the full ACR O-RADS US guideline content, including lexicon definitions and scoring rules, into the prompt. Models were instructed to directly output the final O-RADS category.

**Feature-based hybrid method:** To improve interpretability and stability, we implemented a two-stage hybrid framework. In Stage 1, the LLM served as a structured information extractor, converting unstructured reports into standardized JSON outputs containing O-RADS–related features. To ensure high extraction accuracy, the prompt instructions were iteratively refined using a subset of development reports; the finalized prompt template is provided in Appendix S1. The extraction encompassed 17 core descriptors, which were categorized into (1) classic benign lesion types and (2) morphological and vascular features (detailed in Table 1). Within this schema, papillary projections were treated as count variables, flow score as an ordinal variable (1–4), and all remaining descriptors as binary variables.

In Stage 2, the structured features extracted by the LLM in Stage 1 served as input to a deterministic Python-based rule engine. This rule engine applied the ACR O-RADS US decision logic to assign the final O-RADS category. To support transparent interpretation, the extracted features were output alongside the final O-RADS category.

## 2.2. Dataset Collection and Preprocessing

This retrospective study was approved by the Ethics Committee of Qilu Hospital of Shandong University (No. KYLL-202502-033). The requirement for informed consent was waived due to its retrospective design. The patient selection workflow is depicted in Fig. 2. Ultrasound reports were retrospectively collected from the institutional radiology information system at Qilu Hospital of Shandong University from January 2025 to June 2025. All ultrasound reports were written in Chinese. Eligible cases included female patients who underwent transvaginal, transrectal, or transabdominal pelvic ultrasound examinations with reports describing at least one adnexal lesion. Reports were excluded if they described only physiological adnexal findings (e.g., ovarian follicles or corpus luteum), contained incomplete or missing descriptions of key imaging features, or lacked an assigned O-RADS category.

After applying these criteria, 404 adnexal lesions from 319 patients were included in the final dataset. The dataset was partitioned at the patient level into a development set (n=9 patients; 14 lesions) for prompt optimization and an independent test set (n=310 patients; 390 lesions) for final evaluation. To ensure data privacy and prevent information leakage during model evaluation, all reports were preprocessed using Python scripts followed by manual verification. This process included de-identification by removing all personally identifiable information (e.g., patient names, medical record numbers, and examination dates), as well as removal of original O-RADS classifications to prevent information leakage. Representative examples of the ultrasound reports are provided in Appendix S2.

### 2.3. Reference Standard Establishment

To establish the reference standard, two radiologists (C.Q. and X.L.), each with more than 10 years of experience in gynecologic ultrasonography, independently reviewed all ultrasound reports.

The review process consisted of two stages. First, key sonographic features (e.g., solid components and color Doppler flow scores) were annotated in accordance with the ACR O-RADS lexicon (v2020) [6]. Second, each lesion was assigned an O-RADS category. O-RADS category 1, representing normal or physiologic findings, was not included in our analyses, as the study focused exclusively on non-physiologic adnexal lesions.

To ensure that classification relied solely on sonographic features as described in the ultrasound reports, the readers were blinded to all clinical information, including tumor markers (e.g., CA125 and HE4) and histopathological outcomes. Inter-reader agreement for O-RADS classification was 93.8% (379 of 404 lesions). Discrepancies were resolved through consensus with a third senior radiologist (G.T., with 30 years of experience), and the final consensus served as the reference standard. All analyses were performed at the lesion level. For patients with multiple adnexal lesions, each lesion was treated as a separate analytical unit and assigned an individual reference O-RADS category. For reports describing multiple lesions, lesion location and size were used to match model-generated outputs to the corresponding reference lesions. If a model failed to identify a target lesion, that lesion was counted as a missed lesion and classified as an incorrect prediction.

## 2.4. LLMs and Inference Settings

Eight representative LLMs were evaluated, representing diverse architectures and model capacities, including GPT-5.5 and GPT-5.6 Sol (OpenAI), Claude Fable 5 (Anthropic), Gemini 3.1 Pro and Gemini 3.6 Flash (Google DeepMind), Grok 4.5 (xAI), DeepSeek-V4-Pro (DeepSeek-AI), and Qwen3.7-Max (Alibaba Cloud). All models were accessed through their official application programming interfaces (APIs). Although several models support multimodal inputs, this study focused exclusively on natural language understanding and reasoning based on unstructured clinical text. To enhance reproducibility and reduce output variability, the temperature parameter was fixed at 0 for all models, while all other inference hyperparameters (e.g., top-p and frequency penalty) were retained at the default values specified by the respective APIs.

### 2.5. Statistical Analyses

Statistical analyses were performed in Python (v3.9.12), primarily utilizing the *scikit-learn* and *scipy.stats* libraries. O-RADS categories assigned in the original reports were treated as the human baseline, and expert consensus as the reference standard.

Overall accuracy was the primary metric for O-RADS classification, supplemented by stratified accuracy for each category (O-RADS 2–5). For feature extraction (Stage 1 of the hybrid method), performance was quantified using accuracy, precision, and recall. The McNemar test was used to compare paired classification performance differences (two-sided $P < 0.05$).

Agreement was assessed using quadratic weighted $\kappa$ coefficients with 95% confidence intervals (CIs), which account for the ordinal nature of O-RADS categories and penalize larger category discrepancies. The resulting $\kappa$ values were interpreted as follows: 0.00–0.20, slight; 0.21–0.40, fair; 0.41–0.60, moderate; 0.61–0.80, substantial; and 0.81–1.00, almost perfect [29].

The computer code and pipeline used for data analysis and modeling in this study are publicly available on GitHub at: https://github.com/Xiaotong66/ORADS-LLM-Pipeline.

## 3. Results

### 3.1. Dataset Characteristics

Of the 319 eligible patients, a subset of nine pelvic US reports (corresponding to 14 lesions in 9 female patients; median age, 43 years; range, 34–71 years) was selected for prompt engineering and the construction of in-context examples. These cases were excluded from the subsequent performance evaluation.

The final test set comprised 310 US reports (corresponding to 390 lesions in 310 female patients; median age, 50 years; range, 5–92 years). Detailed characteristics of the lesions are summarized in Table 1.

### 3.2. Comparison of O-RADS Classification Strategies

Three reasoning strategies were evaluated across multiple LLMs: implicit-knowledge end-to-end, rule-informed end-to-end, and feature-based hybrid methods. Overall accuracy for each model–strategy combination is shown in Fig. 3, and category-stratified accuracy is summarized in Table 2.

When relying solely on the model's internal implicit knowledge for direct prediction, overall accuracy ranged from 65.6% to 92.8% (Table 2), with significant performance variations among models. GPT-5.5 achieved the highest overall accuracy under this strategy (92.8%), followed by GPT-5.6 Sol (91.3%) and Gemini 3.6 Flash (90.5%). However, substantial model-to-model variability remained, particularly for O-RADS 3, for which accuracy ranged from 34.4% to 86.9%.

Incorporating explicit O-RADS diagnostic rules (the rule-informed end-to-end strategy) significantly enhanced the stability and consistency of model performance. The overall accuracy of all models improved notably compared with the implicit paradigm, ranging from 82.6% to 95.9%. GPT-5.6 Sol achieved the highest overall accuracy under this strategy (95.9%), and all models except Claude Fable 5 surpassed the original-report baseline of 87.7%. However, performance remained variable for intermediate-risk categories (e.g., O-RADS 3) in some models.

In contrast, the feature-based hybrid method demonstrated the most consistent classification performance across models, with overall accuracy ranging from 95.6% to 99.2%. All hybrid models exceeded the original-report baseline, and the hybrid strategy showed higher and more stable performance overall than the two end-to-end strategies. Three models—Grok 4.5, GPT-5.5, and GPT-5.6 Sol—achieved an identical high overall accuracy of 98.7%, exhibiting exceptionally balanced and robust performance across all individual risk categories. Notably, Gemini 3.6 Flash achieved the highest overall accuracy of 99.2% (387 of 390 lesions), including 100.0% accuracy in the O-RADS 4 category.

Confusion matrix analysis for Gemini 3.6 Flash (Fig. 4) showed that the feature-based hybrid method markedly reduced the number of misclassifications, particularly for O-RADS 2 and O-RADS 3 lesions, compared with both end-to-end strategies and the original clinical reports. Furthermore, it **reduced multi-category misclassifications**, including errors such as classifying an O-RADS 2 lesion as O-RADS 4, which were observed in the original clinical reports.

### 3.3. Agreement between Original Report, Models, and the Reference Standard

For the agreement analysis (Table S1), we selected three representative high-performing models from different model families: Gemini 3.6 Flash, Grok 4.5, and GPT-5.6 Sol. These models achieved the highest overall accuracies with the feature-based hybrid method. The original O-RADS scoring in clinical reports demonstrated almost perfect agreement with the reference standard ($\kappa$ = 0.94; 95% CI: 0.91–0.96). Transitioning to the feature-based hybrid method maximized the performance of all three models, consistently surpassing the performance of original clinical reports.

Although the selected high-performing models already showed almost perfect agreement with the reference standard under the implicit-knowledge end-to-end strategy (weighted κ range, 0.95–0.97), incorporating O-RADS rules further improved agreement under the rule-informed strategy (weighted κ range, 0.96–0.98). The feature-based hybrid method achieved the highest agreement for all three representative models, with weighted κ values of 1.00 for Gemini 3.6 Flash, 0.99 for GPT-5.6 Sol, and 0.99 for Grok 4.5. All three hybrid-model κ values exceeded the original-report baseline. This pattern supports the advantage of decoupling free-text feature extraction from deterministic O-RADS rule execution.

## 3.4. Performance of the Hybrid Model in Feature Classification

To evaluate how the underlying foundation model impacts the hybrid architecture, we compared Gemini 3.6 Flash and Claude Fable 5—the models that yielded the highest and lowest overall accuracies, respectively, within the hybrid framework. Table S2 details their performance across 16 key O-RADS features. Overall, the feature classification accuracy of Gemini 3.6 Flash was significantly higher than that of Claude Fable 5 (98.9% vs 97.8%; $P < 0.001$). Gemini 3.6 Flash also showed higher overall recall (98.3% vs 96.8%) and precision (90.7% vs 82.6%).

At the specific feature level, both models exhibited high consistency in extracting explicitly stated variables, that is, features directly described in the report that require only basic information extraction rather than complex medical inference. They achieved perfect or near-perfect metrics (100%) for classic benign diagnoses (e.g., hemorrhagic cyst, paraovarian cyst) and maintained high accuracy (99.2 - 99.5%) for quantitative or ordinal variables (e.g., papillary projections, color score). However, results for these benign diagnoses should be interpreted cautiously due to their extremely low prevalence in the test set ($n < 10$).

The performance gap between the two models widened when identifying features that require complex semantic interpretation. Specifically, for the extraction of "ascites and/or peritoneal nodules", Gemini 3.6 Flash significantly outperformed Claude Fable 5 in accuracy (99.0% vs 95.1%; $P < 0.001$), driven by a substantially higher precision (98.5% vs 80.0%). Additionally, Gemini 3.6 Flash achieved higher performance in identifying endometriomas (99.2% vs 97.4%) and unilocular cysts (90.5% vs 80.5%).

### 3.5. Analysis of Misclassification Errors

Before analyzing specific misclassifications, the overall distribution of predicted categories was evaluated (Fig. S1). Compared with the original clinical reports and rule-informed strategies, the feature-based hybrid method showed improved alignment with the reference standard.

Misclassifications were categorized by direction and magnitude of deviation from the reference standard (Fig. 5), and mapped onto two primary dimensions: missed diagnosis or understaging on the left, shown in gray and red shades, and overstaging on the right, shown in blue shades.

Original clinical reports exhibited a markedly asymmetrical distribution, with errors predominantly concentrated in the overstaging direction, characterized mainly by mild overstaging and severe overstaging, with few instances of mild understaging.

For the rule-informed end-to-end strategy (Fig. 5A), the total number and direction of misclassifications varied across models. Claude Fable 5 showed the largest number of errors, whereas GPT-5.6 Sol showed the fewest errors. Several models demonstrated a mixed pattern of understaging and overstaging, rather than the predominantly overstaging pattern observed in the original clinical reports. Within the feature-based hybrid framework (Fig. 5B), the total number of misclassifications was markedly reduced across all models, particularly with respect to overstaging. Gemini 3.6 Flash showed the fewest errors, followed by Grok 4.5, GPT-5.5, and GPT-5.6 Sol.

Despite the superior overall performance of the Gemini 3.6 Flash hybrid model, three classification errors were observed. These included one feature hallucination and two feature extraction omissions. Specifically, one O-RADS 2 lesion was classified as O-RADS 1 because of hallucinated physiologic cyst features; one O-RADS 3 lesion was classified as O-RADS 2 because irregular morphology was not extracted; and one O-RADS 5 lesion was classified as O-RADS 3 because ascites-related information was not correctly identified. Descriptions of the relevant reports and involved imaging features for all misclassified cases are summarized in Table S3.

## 4. Discussion

While LLMs excel at processing clinical texts, their accuracy in automating multi-step risk stratification frameworks, such as the ultrasound-based O-RADS, requires further validation. In this study, we systematically compared three LLM-based reasoning strategies to evaluate their reliability in automated O-RADS US classification of ovarian masses. The primary finding was that the feature-based hybrid method achieved an overall accuracy ranging from 95.6% to 99.2%. Notably, when employing the more advanced Gemini 3.6 Flash model, the accuracy peaked at 99.2% (387 of 390 lesions). This hybrid Gemini 3.6 Flash approach demonstrated almost perfect agreement with the reference standard (weighted $\kappa$ = 1.00; 95% CI: 0.99–1.00), exceeding the original reports (accuracy: 87.7%; weighted $\kappa$ = 0.94) and all evaluated end-to-end LLM strategies (accuracy range: 65.6%–95.9%). These findings suggest that while model capability contributes to performance, reasoning architecture plays a decisive role in achieving reliable guideline-based clinical classification.

Prior studies evaluating LLMs in radiology have primarily focused on tasks such as information extraction from free-text reports or direct diagnostic prediction. For example, Huang et al. [27] reported that GPT-3.5 achieved high accuracy in extracting information from free-text medical notes, and Firoozeh et al. [16] observed substantial agreement ($\kappa$ = 0.89–0.90) between LLMs and radiologists in PI-RADS classification. Our results align with prior evidence that LLMs possess strong medical natural language understanding capabilities. However, our study extends this literature by revealing a critical performance gap: reliance on end-to-end generative reasoning alone is insufficient for the rigorous execution of hierarchical guidelines. In our experiments, even with explicit guideline prompting, rule-informed end-to-end approaches exhibited notable instability (accuracy range: 82.6%–95.9%), suggesting limitations in reliably operationalizing complex decision trees within unconstrained generative workflows. Furthermore, while previous research has often been limited by small model cohorts or a focus on MRI-based O-RADS [23], our study provides a systematic comparison of reasoning strategies across multiple contemporary LLMs specifically for ultrasound-based O-RADS classification.

To overcome these limitations, we implemented an architectural decoupling strategy that separates feature extraction from deterministic rule execution. This framework mirrors the cognitive workflow of radiologists, who first identify descriptive findings before mapping them to standardized risk categories. Consequently, compared with traditional end-to-end strategies, our hybrid framework yielded greater consistency and significantly reduced performance variability. Nevertheless, such structural optimization cannot fully overcome the inherent limitations of the model, as the intrinsic capability of the underlying LLM remains a key determinant of the final performance. For instance, Gemini 3.6 Flash achieved a higher overall feature extraction accuracy than Claude Fable 5 (98.9% vs 97.8%; $P < 0.001$). This advantage was particularly pronounced when identifying complex findings, such as ascites or peritoneal nodules (99.0% vs 95.1%; $P < 0.001$). These feature-level differences help explain the residual variation in downstream O-RADS classification performance across models.

From a clinical safety perspective, the direction of error is as important as overall accuracy. Consistent with prior literature describing a tendency toward overstaging in routine radiologic practice [30], the original clinical reports in our cohort showed a similar pattern. In contrast, certain end-to-end LLM strategies demonstrated unpredictable understaging or overstaging, even when guideline content was explicitly provided. The hybrid method significantly reduced overall misclassifications, mitigated the overstaging tendency observed in the original clinical reports, and provided transparent intermediate outputs to facilitate clinical oversight.

Several limitations should be acknowledged. First, this study was conducted at a single institution, and external validation across diverse clinical settings is warranted to further assess the generalizability of the proposed framework. Differences in ultrasound reporting styles and institutional practices may influence feature extraction performance. Second, certain sonographic descriptors were infrequently represented, introducing some statistical uncertainty in feature-level performance estimates. Finally, this system is intended as a decision-support tool rather than a replacement for radiologists, and its performance depends on the completeness and accuracy of the underlying ultrasound reports.

## 5. Conclusions

For complex, guideline-driven medical classification tasks such as O-RADS, hybrid architectures that explicitly separate feature extraction from rule-based reasoning achieve higher classification accuracy with greater reliability and interpretability than end-to-end LLM approaches. Importantly, our findings suggest that improving LLM-based clinical decision-making may depend less on increasing model complexity alone and more on designing appropriate reasoning architectures that align with clinical workflows. Future research should prioritize prospective, multicenter validation across diverse populations. Exploration of multimodal LLM systems integrating both ultrasound images and corresponding textual reports may further enhance clinical applicability and workflow efficiency.

## Declarations

### Abbreviations

ACR: American College of Radiology

ADNEX: Assessment of Different Neoplasias in the Adnexa

CI: Confidence interval

IOTA: International Ovarian Tumor Analysis

LLM: Large language model

O-RADS: Ovarian-Adnexal Reporting and Data System

US: Ultrasound

### Ethics approval and consent to participate

The study was conducted in accordance with the Declaration of Helsinki and was approved by the Ethics Committee of Qilu Hospital of Shandong University (No. KYLL-202502-033). The requirement for written informed consent was waived by the ethics committee because of the retrospective study design and the use of de-identified clinical data.

### Consent for publication

Not applicable. This manuscript does not contain identifiable individual patient data.

### Availability of data and materials

The datasets generated and/or analyzed during the current study are not publicly available due to patient privacy and institutional restrictions but are available from the corresponding author on reasonable request. The computer code and pipeline used for data analysis and modeling in this study are publicly available on GitHub at: https://github.com/Xiaotong66/ORADS-LLM-Pipeline.

## Competing Interests

The authors declare that they have no competing interests.

## Funding

This work was supported by Guangdong Basic and Applied Basic Research Foundation (Nos. 2025A1515011821 and 2026A1515012488 to JC), Shenzhen Science and Technology Program (No. JCYJ20240813143302004 to JC), National Natural Science Foundation of China (Nos. 12326619 and 62171290 to DN), Science and Technology Planning Project of Guangdong Province (No. 2023A0505020002 to DN), and Frontier Technology Development Program of Jiangsu Province (No. BF2024078 to DN).

## Authors' contributions

X.T. and C.Q. wrote the main manuscript text. X.T. contributed to formal analysis, methodology, and software development. C.Q. contributed to investigation and data curation. X.L. contributed to data curation and validation. Q.H. contributed to methodology. G.Z., B.G., X.S., S.W., and X.W. contributed to data curation. W.X. contributed to validation. R.H. prepared the figures. D.N. contributed to funding acquisition and project administration. G.T. contributed to data curation, resources, supervision, and manuscript review and editing. J.C. contributed to conceptualization, funding acquisition, supervision, and manuscript review and editing. All authors reviewed and approved the final manuscript.

## Acknowledgements

Not applicable.

# Tables

**Table 1. Baseline characteristics of the study population.**

| Characteristic | Training set (n = 14) | Test set (n = 390) |
|---|---|---|
| **Patient demographics** | | |
| Age (y) | 43 (39-55) | 50 (49-55) |
| **Extracted features: classic benign lesions** | | |
| Physiologic cyst | 0 (0.0) | 5 (1.3) |
| Hemorrhagic cyst | 0 (0.0) | 2 (0.5) |
| Dermoid cyst | 1 (7.1) | 52 (13.3) |
| Endometrioma | 0 (0.0) | 52 (13.3) |
| Paraovarian cyst | 0 (0.0) | 7 (1.8) |
| Peritoneal inclusion cyst | 0 (0.0) | 0 (0.0) |
| Hydrosalpinx | 0 (0.0) | 2 (0.5) |
| **Extracted features: morphology & vascularity** | | |
| Unilocular | 5 (35.7) | 74 (19.0) |
| Bilocular | 1 (7.1) | 11 (2.8) |
| Multilocular | 2 (14.3) | 57 (14.6) |
| Solid component | 3 (21.4) | 71 (18.2) |
| Solid lesion | 5 (35.7) | 108 (27.7) |
| Irregular wall/septation | 2 (14.3) | 75 (19.2) |
| Acoustic shadowing | 1 (7.1) | 8 (2.1) |
| Ascites and/or peritoneal nodules | 3 (21.4) | 70 (17.9) |
| Papillary projection (≥1) | 0 (0.0) | 35 (9.0) |
| Color score (1-4) | 1.5 (1-2) | 2 (1-3) |
| **O-RADS** | | |
| O-RADS 2 | 4 (28.6) | 145 (37.2) |
| O-RADS 3 | 3 (21.4) | 61 (15.6) |
| O-RADS 4 | 2 (14.3) | 63 (16.2) |
| O-RADS 5 | 5 (35.7) | 121 (31.0) |

**Note.** Unless otherwise indicated, continuous variables are medians with IQRs in parentheses; categorical variables are numbers with percentages in parentheses. Papillary projections, although extracted as counts, are reported as categorical data (presence/absence) due to the predominance of zero values. All characteristics are presented on a per-lesion basis.

**Table 2. Accuracy (%) of three LLM-based strategies for automated O-RADS classification stratified by risk category.**

| Model | O-RADS 2 | O-RADS 3 | O-RADS 4 | O-RADS 5 | Overall |
|---|---|---|---|---|---|
| **Original report** | 76.6 | 88.5 | 90.5 | 99.2 | 87.7 |
| **Implicit-knowledge end-to-end** | | | | | |
| Claude Fable 5 | 80.7 | 34.4 | 68.3 | 62.0 | 65.6 |
| DeepSeek-V4-Pro | 88.3 | 57.4 | 85.7 | 73.6 | 78.5 |
| Qwen3.7-Max | 93.8 | 55.7 | 77.8 | 92.6 | 84.9 |
| Grok 4.5 | 90.3 | 78.7 | 85.7 | 88.4 | 87.2 |
| Gemini 3.1 Pro | 84.8 | 75.4 | 92.1 | 98.3 | 88.7 |
| Gemini 3.6 Flash | 86.2 | 86.9 | 95.2 | 95.0 | 90.5 |
| GPT-5.5 | 95.9 | 83.6 | 98.4 | 90.9 | **92.8** |
| GPT-5.6 Sol | 95.2 | 78.7 | 100.0 | 88.4 | 91.3 |
| **Rule-informed end-to-end** | | | | | |
| Claude Fable 5 | 89.7 | 60.7 | 74.6 | 89.3 | 82.6 |
| DeepSeek-V4-Pro | 93.8 | 82.0 | 88.9 | 90.1 | 90.0 |
| Qwen3.7-Max | 92.4 | 85.2 | 85.7 | 94.2 | 90.8 |
| Grok 4.5 | 95.9 | 88.5 | 98.4 | 94.2 | 94.6 |
| Gemini 3.1 Pro | 95.2 | 83.6 | 100.0 | 92.6 | 93.3 |
| Gemini 3.6 Flash | 92.4 | 85.2 | 98.4 | 95.0 | 93.1 |
| GPT-5.5 | 94.5 | 93.4 | 98.4 | 91.7 | 94.1 |
| GPT-5.6 Sol | 94.5 | 96.7 | 100.0 | 95.0 | **95.9** |
| **Feature-based hybrid method** | | | | | |
| Claude Fable 5 | 96.6 | 91.8 | 92.1 | 98.3 | 95.6 |
| DeepSeek-V4-Pro | 97.9 | 100.0 | 95.2 | 99.2 | 98.2 |
| Qwen3.7-Max | 98.6 | 98.4 | 96.8 | 98.3 | 98.2 |
| Grok 4.5 | 97.9 | 98.4 | 98.4 | 100.0 | 98.7 |
| Gemini 3.1 Pro | 100.0 | 95.1 | 98.4 | 97.5 | 98.2 |
| Gemini 3.6 Flash | 99.3 | 98.4 | 100.0 | 99.2 | **99.2** |
| GPT-5.5 | 99.3 | 98.4 | 98.4 | 98.3 | 98.7 |
| GPT-5.6 Sol | 99.3 | 98.4 | 98.4 | 98.3 | 98.7 |

**Note.** Data are percentages of lesions correctly classified compared with the reference standard. The “Original report” represents the baseline performance of the original clinical ultrasound reports. The "Implicit-knowledge" paradigm relies on the model's internal parameters; the "Rule-informed" paradigm incorporates ACR O-RADS guidelines into the prompt; and the "Feature-based hybrid" paradigm utilizes a two-stage extract-then-reason architecture. Bold indicates the highest overall accuracy within each reasoning strategy.

# Figures

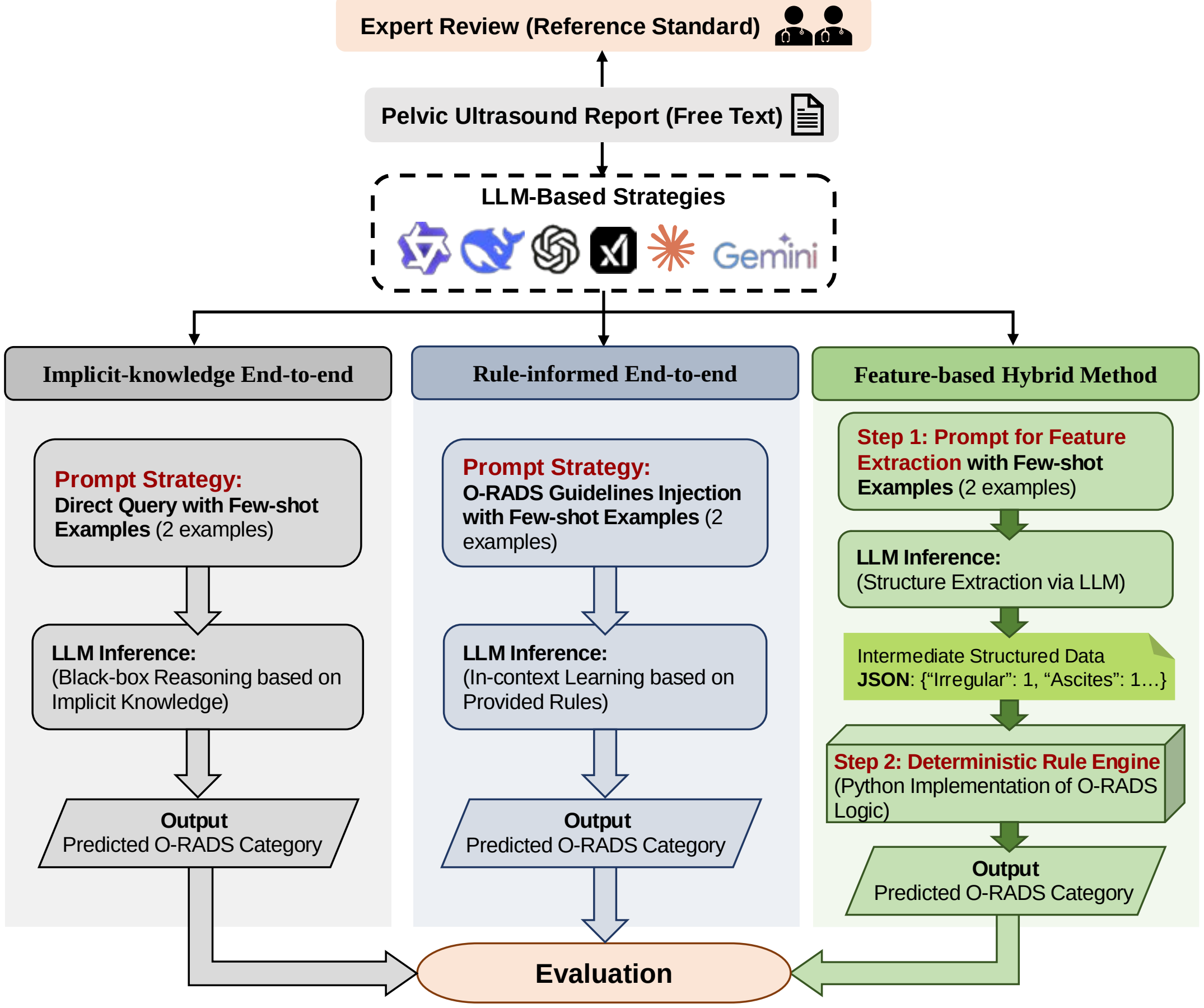

**Figure 1. Overview of the experimental workflow comparing three LLM-based strategies for O-RADS classification.** The workflow compares: (left) the Implicit-knowledge End-to-end Strategy, which relies on direct few-shot querying; (middle) the Rule-informed End-to-end Strategy, which incorporates O-RADS guidelines into the prompt; and (right) the Feature-based Hybrid Method, a two-step strategy that decouples LLM-based feature extraction from deterministic rule-based scoring.

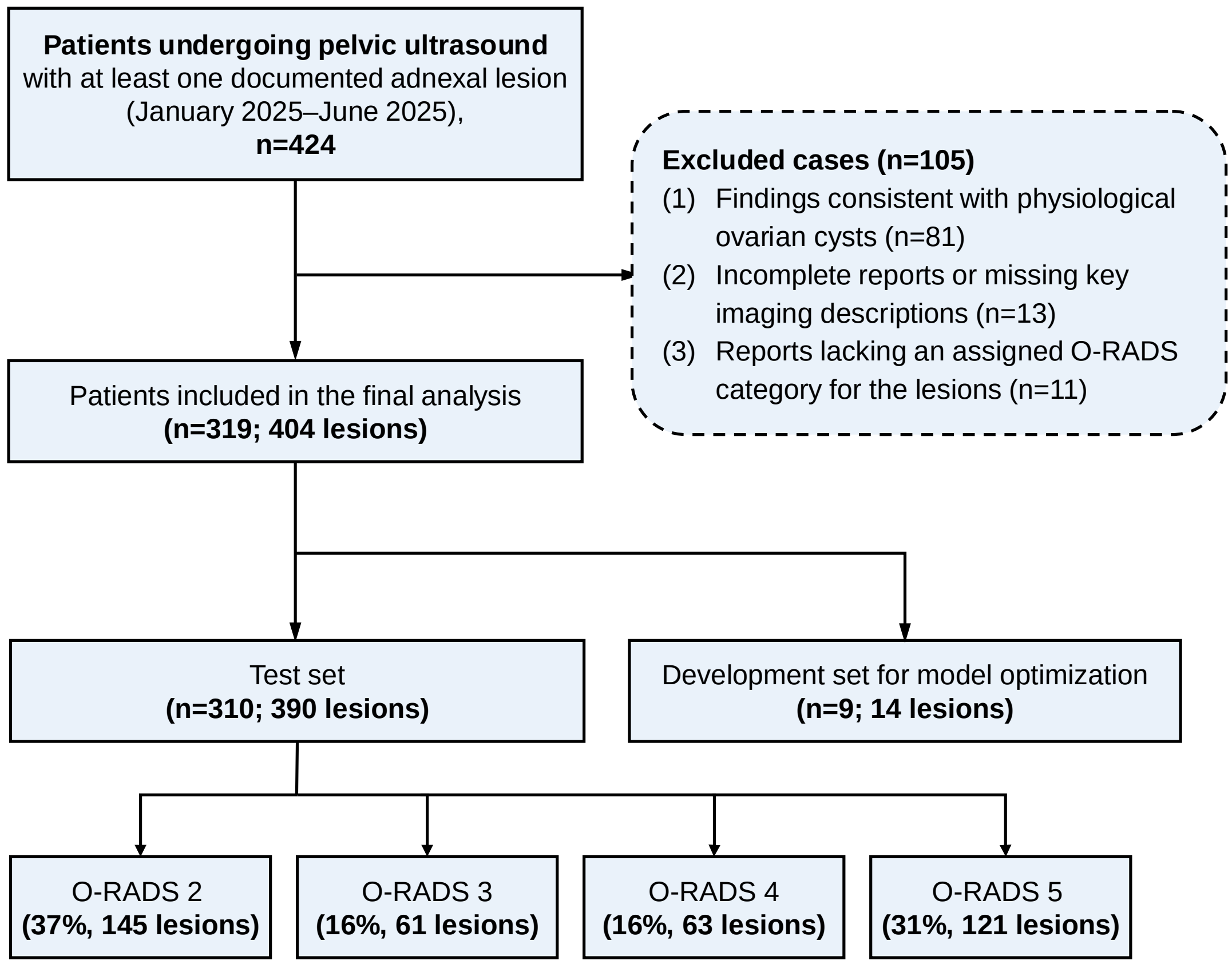

**Figure 2. Flowchart of patient selection and dataset construction.** A total of 424 patients with at least one documented adnexal lesion between January 2025 and June 2025 were screened. After applying exclusion criteria, 319 patients (404 lesions) were included in the final cohort. The dataset was divided into a development set for model optimization (n = 9; 14 lesions) and a test set for evaluation (n = 310; 390 lesions), with distribution across O-RADS categories shown.

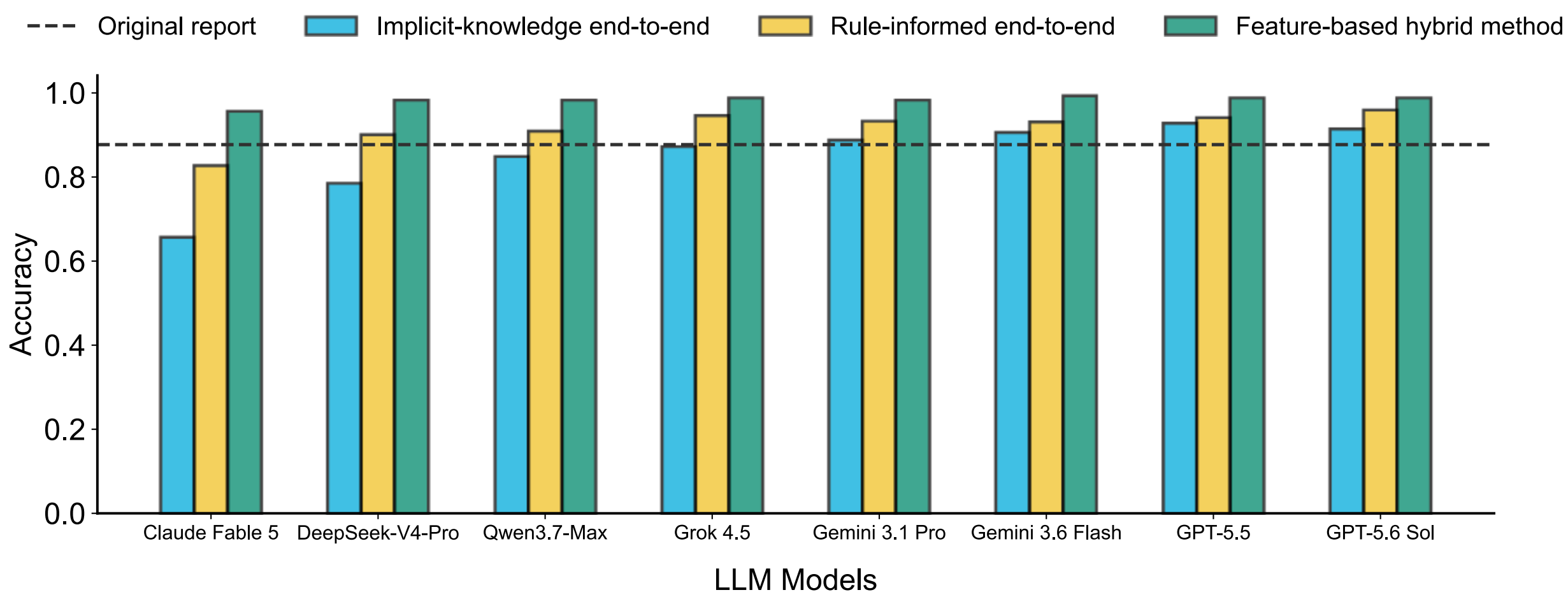

**Figure 3. Overall classification accuracy of eight LLMs across three strategies.** The bar chart compares the performance of Implicit-knowledge End-to-end (blue), Rule-informed End-to-end (yellow), and Feature-based Hybrid (green) strategies. The horizontal dashed line represents the baseline classification accuracy of the original clinical ultrasound reports.

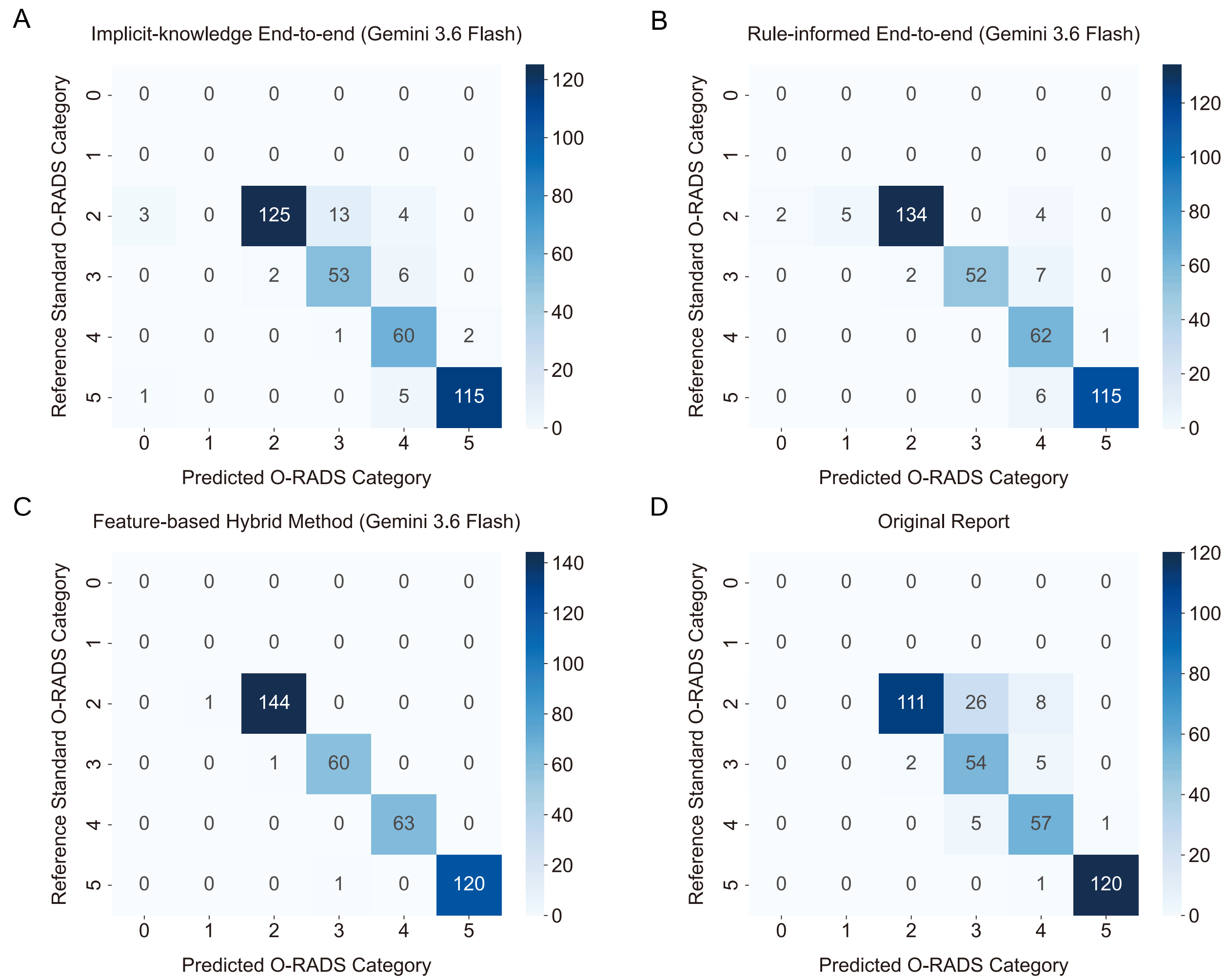

**Figure 4. Confusion matrices comparing O-RADS classification performance for Gemini 3.6 Flash across three reasoning strategies and the original clinical report.** (A) Implicit-knowledge End-to-end strategy. (B) Rule-informed End-to-end strategy. (C) Feature-based Hybrid Method. (D) Original Report. The x-axis represents the predicted O-RADS category, and the y-axis represents the reference standard. Values on the diagonal indicate correctly classified lesions, whereas off-diagonal values represent misclassifications. The Feature-based Hybrid Method (C) showed the fewest off-diagonal misclassifications.

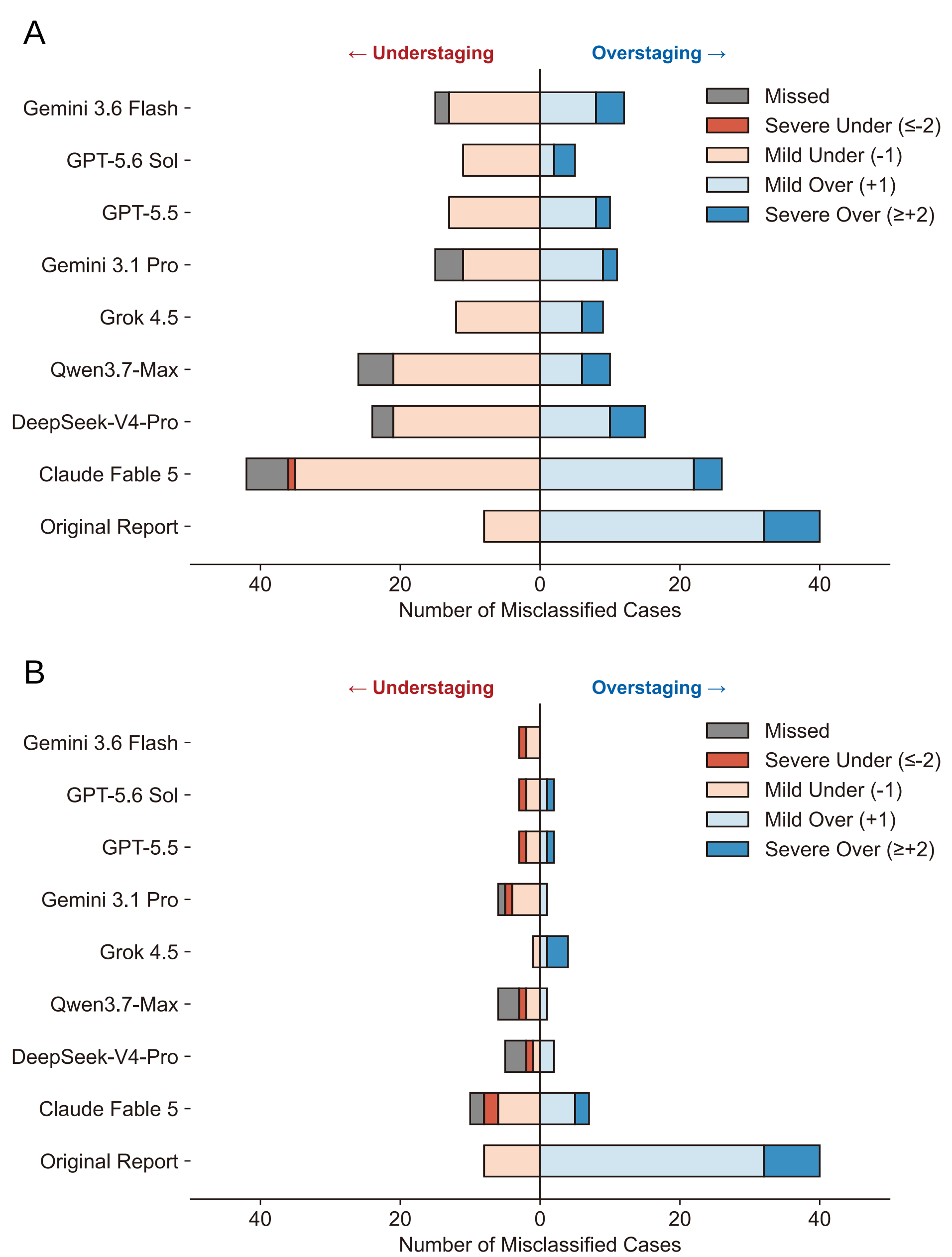

**Figure 5. Stratified analysis of misclassification errors comparing the (A) Rule-informed End-to-end Strategy and (B) Feature-based Hybrid Method with the Original Report.** The diverging stacked bar charts illustrate the clinical direction and severity of errors relative to the reference standard. The vertical axis separates missed diagnosis or understaging (left; representing risk of missed diagnosis or undertreatment) from overstaging (right; representing risk of overtreatment).

# Supplemental Material

# Improving O-RADS Risk Stratification from Ultrasound Reports: A Comparative Evaluation of Hybrid versus End-to-End LLM Reasoning Strategies

**Contents**

## Supplementary Tables

### Table S1. Pairwise agreement of O-RADS categories.

| Comparison pair | Weighted $\kappa$ value |
|---|---|
| **Baseline comparison** | |
| Original report vs reference standard | 0.94 (0.91–0.96) |
| **Gemini 3.6 Flash vs reference standard** | |
| Implicit-knowledge end-to-end | 0.95 (0.91–0.97) |
| Rule-informed end-to-end | 0.96 (0.95–0.98) |
| Feature-based hybrid method | 1.00 (0.99–1.00) |
| **GPT-5.6 Sol vs reference standard** | |
| Implicit-knowledge end-to-end | 0.97 (0.95–0.98) |
| Rule-informed end-to-end | 0.98 (0.97–0.99) |
| Feature-based hybrid method | 0.99 (0.98–1.00) |
| **Grok 4.5 vs reference standard** | |
| Implicit-knowledge end-to-end | 0.95 (0.94–0.97) |
| Rule-informed end-to-end | 0.98 (0.96–0.99) |
| Feature-based hybrid method | 0.99 (0.98–1.00) |

**Note.** Data in parentheses are 95% CIs. Quadratic weighted kappa coefficients were employed to penalize larger disagreements between O-RADS categories.

**Table S2. Performance of Claude Fable 5 and Gemini 3.6 Flash for O-RADS feature extraction compared with the reference standard review.**

| O-RADS feature | Accuracy (%) | | | Recall (%) | | Precision (%) | |
|---|---|---|---|---|---|---|---|
| | Claude Fable 5 | Gemini 3.6 Flash | *P* value | Claude Fable 5 | Gemini 3.6 Flash | Claude Fable 5 | Gemini 3.6 Flash |
| **Overall (n=390)** | 97.8 | 98.9 | <.001 | 96.8 | 98.3 | 82.6 | 90.7 |
| Physiologic cyst (n=5) * | 99.7 | 99.7 | >.99 | 80.0 | 100.0 | 100.0 | 83.3 |
| Hemorrhagic cyst (n=2) * | 100.0 | 100.0 | >.99 | 100.0 | 100.0 | 100.0 | 100.0 |
| Dermoid cyst (n=52) | 99.0 | 100.0 | .13 | 96.2 | 100.0 | 96.2 | 100.0 |
| Endometrioma (n=52) | 97.4 | 99.2 | .04 | 94.2 | 98.1 | 87.5 | 96.2 |
| Paraovarian cyst (n=7) * | 100.0 | 100.0 | >.99 | 100.0 | 100.0 | 100.0 | 100.0 |
| Hydrosalpinx (n=2) * | 100.0 | 99.7 | >.99 | 100.0 | 100.0 | 100.0 | 66.7 |
| Unilocular (n=74) | 80.5 | 90.5 | <.001 | 95.9 | 98.6 | 49.3 | 67.0 |
| Bilocular (n=11) | 100.0 | 99.7 | >.99 | 100.0 | 100.0 | 100.0 | 91.7 |
| Multilocular (n=57) | 98.2 | 99.0 | .25 | 96.5 | 100.0 | 91.7 | 93.4 |
| Solid component (n=71) | 98.5 | 98.7 | >.99 | 97.2 | 98.6 | 94.5 | 94.6 |
| Solid lesion (n=108) | 99.5 | 99.7 | >.99 | 99.1 | 99.1 | 99.1 | 100.0 |
| Irregular wall/septation (n=75) | 97.2 | 97.7 | .63 | 96.0 | 96.0 | 90.0 | 92.3 |
| Acoustic shadowing (n=8) * | 99.0 | 99.0 | >.99 | 100.0 | 100.0 | 66.7 | 66.7 |
| Ascites and/or peritoneal nodules (n=70) | 95.1 | 99.0 | <.001 | 97.1 | 95.7 | 80.0 | 98.5 |
| Papillary projection † | 99.2 | 99.2 | >.99 | — | — | — | — |
| Color score † | 99.5 | 99.5 | >.99 | — | — | — | — |

**Note**. Data are percentages. *P* values were calculated using the McNemar test for paired comparisons of feature-level classification accuracy between Claude Fable 5 and Gemini 3.6 Flash on the same lesions (n = 390). Overall metrics represent micro-averaged performance across all evaluated features. Peritoneal inclusion cyst was excluded from feature-specific analyses because no positive cases were present in the dataset. * Features with fewer than 10 positive cases should be interpreted with caution due to limited sample size. † Recall and precision were not calculated for papillary projections and color score, as these variables were treated as count or ordinal measures and evaluated using exact category agreement rather than binary classification.

**Table S3. Analysis of all misclassification errors by the hybrid model (Gemini 3.6 Flash).**

| Error Category | Case ID | Report-related Description | Implicated Feature | Clinical Consequence |
|---|---|---|---|---|
| Feature Hallucination | 3-23-1 | Left ovary: The left ovary measures 3.3 × 1.9 cm. A 1.0 × 0.9 cm slightly hyperechoic nodular lesion is seen within the left ovary, with relatively well-defined margins. No obvious internal vascularity is detected on Doppler imaging. | Physiologic cyst | Ref: 2 –> Pred: 1 |
| Feature Extraction Omission | 3-7-0 | Left adnexa: A 7.2 × 5.2 × 5.3 cm unilocular cystic lesion is seen in the left adnexal region, with mildly irregular morphology. No obvious internal vascularity is detected on Doppler imaging. Internal acoustic transmission is suboptimal, with low-level internal echoes/debris. | Irregular | Ref: 3 –> Pred: 2 |
| | 5-62-0 | A patchy anechoic fluid area measuring approximately 4.8 × 2.5 cm was noted adjacent to the lesion, containing reticular internal structures. | Ascites | Ref: 5 –> Pred: 3 |

**Note.** The "Report-related Description" column presents literal English translations of excerpts from the original Chinese ultrasound reports; however, all LLM inference experiments were conducted entirely using the original Chinese text. Feature hallucination refers to instances where the model incorrectly extracted features not supported by the text. Feature extraction omission refers to the failure to identify key features present in the text. *Pred = predicted O-RADS category by the Gemini 3.6 Flash hybrid model.*

## Supplementary Figures

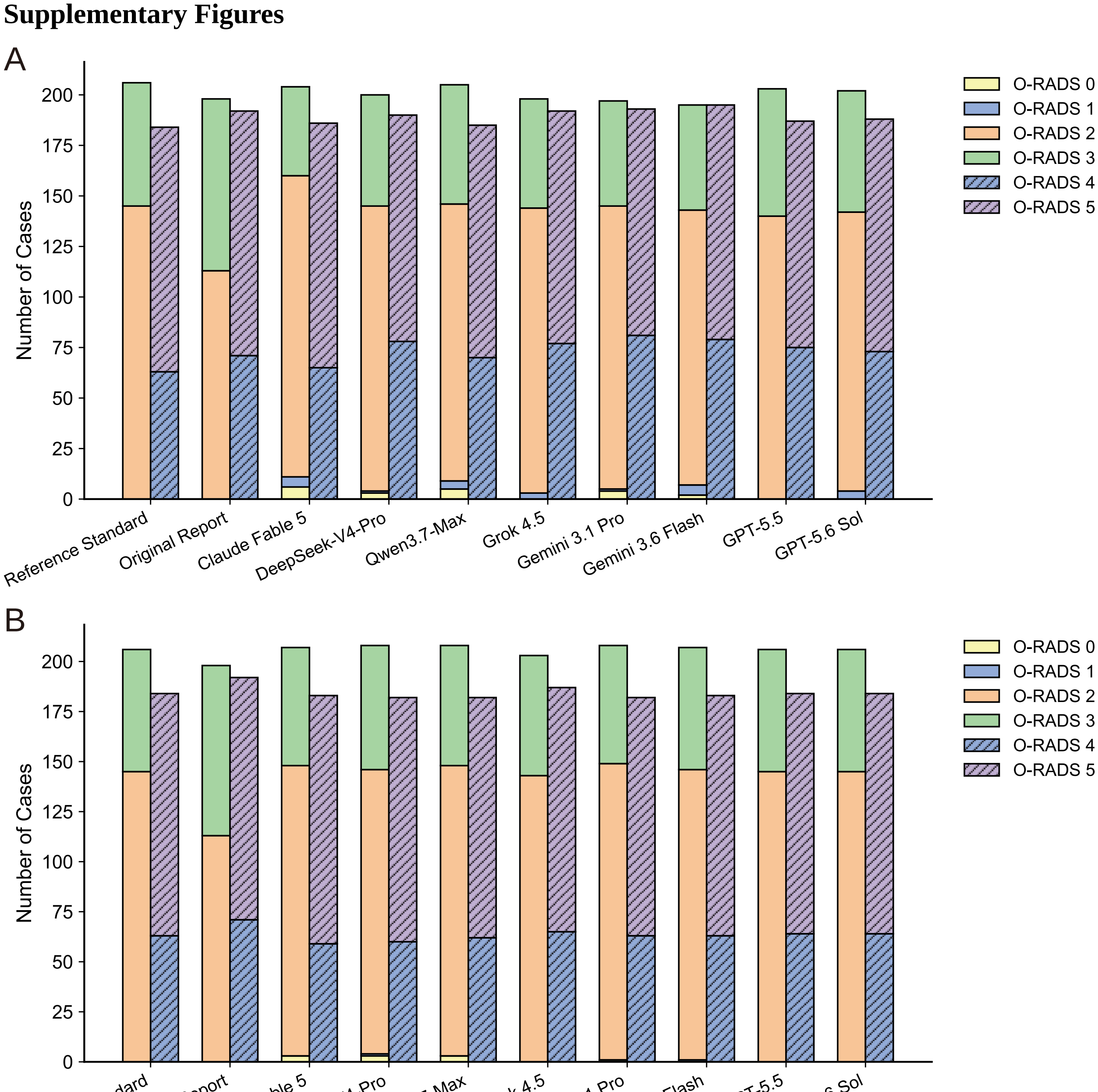

**Figure S1. Overall distribution of predicted O-RADS categories across models and reasoning strategies.** Stacked bar charts illustrate the number of cases assigned to each O-RADS category by the large language models, the reference standard, and the original clinical reports. (A) Category distribution under the rule-informed end-to-end strategy. (B) Category distribution under the feature-based hybrid method. The presence of the “O-RADS 0” category (yellow bars) represents cases in which the models failed to detect or omitted the target ovarian lesions described in the ultrasound reports.

## Supplementary Appendices

## Appendix S1. Prompt template for the hybrid method

Note: All LLM feature extraction and reasoning experiments were conducted entirely using the original Chinese prompts to process the Chinese ultrasound reports. The English translation is provided solely for the reviewers' and international readers' reference to ensure methodological transparency.

### Prompt (English translation)

You are a gynecologic imaging expert. Your task is to extract structured imaging features of ovarian and adnexal lesions based *solely* on the explicit textual descriptions provided in the ultrasound reports, and output the results in a strict JSON format on a per-lesion basis.

Only analyze lesions originating from the ovary, adnexa, or fallopian tube. All fields must be populated based *strictly* on explicit descriptions in the text; any form of inference, association, or supplementation with medical common sense is strictly prohibited. Any feature not explicitly mentioned must be assigned a value of 0. Please note that the output must be a strict JSON format. Do not include markdown code blocks (```), comments, or any explanatory text.

- **Lesion Location:** e.g., output "left adnexa", "right adnexa", "left ovary", or "right ovary".
- **Max Diameter:** Output only the numerical value of the maximum lesion diameter; do not include units.
- **Unilocular cyst:** If the lesion is described as unilocular, cystic echo without septations, mark as 1; otherwise, 0.
- **Bilocular cyst:** If the lesion is described as bilocular, containing a septation, or "one septation seen," mark as 1; otherwise, 0.
- **Multilocular cyst:** If the lesion is described as multilocular, containing multiple septations, several local septations, dense septations, or multiple septations, mark as 1; otherwise, 0.
- **Color Doppler Flow Score:** * If described as "no obvious blood flow signals" or "no blood flow,"

mark as 1.If described as "a small amount of blood flow signals," "punctate blood flow signals," or "sparse blood flow signals," mark as 2.If described as "multiple linear blood flow signals," "linear blood flow," or "moderate blood flow signals," mark as 3.If described as "relatively abundant blood flow signals" or "abundant blood flow," mark as 4.If there is no mention of blood flow, mark as 0.

- **Physiologic cyst:** If the lesion is described as a follicle, corpus luteum, or physiologic cyst, mark as 1; otherwise, 0.
- **Hemorrhagic cyst:** If the lesion is described as a hemorrhagic cyst, mark as 1; otherwise, 0.
- **Dermoid cyst / Teratoma:** If the lesion is described as having "punctate/short linear hyperechoic foci, hyperechoic components, patchy hyperechoic areas" AND has no blood flow AND has < 3 locules; OR if the ultrasound impression suggests a dermoid cyst or teratoma, mark as 1; otherwise, 0.
- **Endometrioma:** If the lesion is described as having "poor acoustic transmission, filled with fine internal echoes" AND has no blood flow AND has < 3 locules; OR if the ultrasound impression suggests a chocolate cyst or endometrioma, mark as 1; otherwise, 0.
- **Paraovarian cyst:** If the lesion is described as an extraovarian unilocular cyst, paraovarian cyst, or the ultrasound impression suggests a paraovarian cyst, mark as 1; otherwise, 0.
- **Peritoneal Inclusion Cyst:** If the lesion is described as an encapsulated fluid collection, or the ultrasound impression suggests a peritoneal inclusion cyst, mark as 1; otherwise, 0.
- **Hydrosalpinx:** If the lesion is described as a tortuous tubular echo, tubular appearance, or the ultrasound impression suggests a hydrosalpinx, mark as 1; otherwise, 0.
- **Solid lesion:** Refers to a lesion where solid components constitute >80% of the total area. If the text explicitly states "predominantly solid," "solid lesion," or "predominantly solid echogenicity," mark as 1; otherwise, 0. *Note: Dermoid cysts, endometriomas, hemorrhagic cysts, and proteinaceous or hyperechoic contents must NOT be counted as solid components.*
- **Solid component:** Refers to the presence of solid components constituting <80% of the lesion. This feature is evaluated ONLY if "Solid lesion" is marked as 0. If the lesion is described as "cystic-solid," "mixed-echogenic mass," or having "solid projections" indicating solid structures, mark as 1; otherwise, 0. *Similarly, dermoid cysts, endometriomas, hemorrhagic cysts, and proteinaceous/hyperechoic contents must NOT be counted as solid components.*

- **Irregular wall / septation:** Refers to an irregular outer wall, inner wall, or septation of a cystic lesion; or an irregular outer contour, irregular outer margin, or irregular shape of a solid lesion. Mark as 1 ONLY if explicitly described with one of the above terms; otherwise, 0.
- **Acoustic shadowing:** If the lesion is described as being accompanied by acoustic shadowing or significant posterior acoustic attenuation, mark as 1; otherwise, 0.
- **Papillary projection:** If there are nodules protruding from the cyst wall into the cavity, irregular papillary projections on the cyst wall, papillary projections, intracystic nodules, or small protrusions on the cyst cavity wall, input the **exact count (number)**; if not explicitly mentioned, output 0.
- **Ascites and/or peritoneal nodules:** If "nodules," "thickening (>0.5 cm)," "ill-defined margins," or "involvement" are mentioned in the peritoneum, pelvic floor, or pouch of Douglas, mark as 1.
  - **(1)** If an anechoic fluid collection is seen in the "perihepatic, perirenal, or bilateral iliac fossae," regardless of depth, consider it ascites and mark as 1.
  - **(2)** If an anechoic fluid collection is detected in the "pelvic and abdominal cavity" or "abdominal cavity," the depth must be > 2.0 cm to be considered ascites and marked as 1; otherwise, it is not.
  - **(3)** If an anechoic fluid collection is detected in other locations (e.g., "pouch of Douglas, pelvic cavity, periuterine, or posterior to the uterus"), the depth must be > 4.5 cm to be considered ascites and marked as 1; otherwise, it is not.
  - **(4)** Note: Ascites, peritoneal nodules, or extensive pelvic implants are global signs. If mentioned in the report, they MUST be marked as 1 for ALL lesion objects currently being output.

If the report contains multiple lesions, you must output a strict JSON array (list), where each element corresponds to a single lesion object. Do not output multiple independent JSON objects, and do not include any explanatory text.

**[Few-Shot Example 1 Input]** Ultrasound Findings: Retroverted uterus, flexed posteriorly due to compression. Uterine body measures 4.5 × 2.1 × 1.9 cm, with heterogeneous myometrial echotexture. Endometrial thickness: 0.16 cm. A mass is detected in the left adnexal region, measuring approximately 16.1 × 15.6 × 9.5 cm. It is predominantly solid with an irregular outer contour, and multiple linear blood flow signals are observed within.

**[Few-Shot Example 1 Extracted Output Fields]** "Lesion Location": "Left adnexal region", "Max Diameter (cm)": 16.1, "Solid lesion": 1, "Irregular wall / septation": 1, "Color Doppler Flow Score": 3

**[Few-Shot Example 2 Input]** Ultrasound Findings: Retroverted uterus, measuring 7.1 × 5.3 × 3.9 cm, with regular morphology and fairly homogeneous myometrial echotexture. Endometrial thickness: 0.8 cm, heterogeneous. Left ovary: 2.5 × 1.7 *cm. A mass measuring approximately* 7.4 × 6.1 cm is seen anterior to the uterine body. It is predominantly solid, accompanied by acoustic shadowing, and sparse blood flow signals are detected within. It is closely adjacent to the left ovary. Right ovary: 2.4 × 1.4 cm. A free anechoic fluid collection is detected in the pouch of Douglas, approximately 3.0 cm in depth, with good acoustic transmission. Ultrasound Impression: Solid mass in the pelvic cavity (suspected thecoma-fibroma group tumor?); pelvic effusion.

**[Few-Shot Example 2 Extracted Output Fields]** "Lesion Location": "Left adnexal region", "Max Diameter (cm)": 7.4, "Solid lesion": 1, "Acoustic shadowing": 1, "Color Doppler Flow Score": 2

**[JSON Template to be Extracted]** {JSON_TEMPLATE}

**[Input Ultrasound Text Report]** {report_text}

## Appendix S2. Representative ultrasound report examples

**Note:** The inference experiments for all LLMs were conducted entirely using the original Chinese ultrasound reports. Prior to inference, all reports were strictly de-identified (removing patient personal information), and any original O-RADS categories or baseline scores were redacted to prevent data leakage. To ensure full

transparency and methodological clarity, we present representative examples of English translations for the reviewers' and readers' reference.

### S2.1 Example 1 (O-RADS 2 Case)

**Ultrasound Findings:** Anteverted uterus, measuring 7.0 × 4.7 × 3.6 cm, with regular morphology and fairly homogeneous myometrial echotexture. Endometrial thickness: 0.75 cm. Left ovary measures 7.2 × 4.6 cm. A mixed-echogenic mass measuring approximately 5.8 × 4.9 × 4.3 cm is detected within it, with well-defined margins. The cystic component demonstrates poor acoustic transmission and is filled with fine internal echoes. The solid component is relatively hyperechoic, measuring approximately 4.0 × 3.2 cm, with no obvious internal blood flow signals detected. Right ovary measures 3.9 × 2.7 cm.

**Ultrasound Impression:** Mixed-echogenic mass in the left ovary (suspected teratoma/dermoid cyst).

### S2.2 Example 2 (O-RADS 3 and O-RADS 4 Case)

**Ultrasound Findings:** Anteverted uterus, measuring 8.4 × 4.5 × 3.6 cm, with regular morphology and fairly homogeneous myometrial echotexture. Endometrial thickness is 1.2 cm. Left ovary measures 3.8 × 1.9 cm, containing a 2.2 × 1.3 cm cystic lesion with sparse internal blood flow signals. Right ovary measures 4.3 × 3.7 cm, containing a 2.6 × 2.0 cm predominantly solid lesion with 1 papillary projection observed on the cyst wall; no obvious blood flow signals are detected within it. Adjacent to this lesion, a 3.3 × 2.4 cm cystic-solid mass is noted, featuring dense septations within the cystic component, along with a small amount of detectable blood flow signals demonstrating an arterial spectral waveform.

**Ultrasound Impression:** 1. Cystic lesion in the right ovary; 2. Cystic-solid mass in the right ovary; 3. Cyst in the left ovary (suspected corpus luteum cyst).

### S2.3 Example 3 (O-RADS 5)

**Ultrasound Findings:** Anteverted uterus, measuring 7.7 × 4.9 × 4.3 cm, with ill-defined margins and heterogeneous myometrial echotexture. Endometrial thickness is 0.24 cm. A predominantly solid, hypoechoic mass is detected posterior and to the right of the uterus, extending to the left of the cervix. The lesions are confluent, measuring approximately 9.5 × 5.4 × 4.4 cm in total extent, with ill-defined margins, irregular shape, and heterogeneous internal echogenicity. Relatively abundant blood flow signals are observed within the mass, and it is inseparable from the posterior uterine wall. Bilateral ovaries are not clearly visualized.

**Ultrasound Impression:** Hypoechoic mass located posterior and to the right of the uterus, and to the left of the cervix.